\documentclass[10pt, a4paper]{article}

\usepackage{lrec-coling2024} 
\usepackage{booktabs}

\usepackage{subfig}
\usepackage[normalem]{ulem}
\useunder{\uline}{\ul}{}

\title{ESG Classification by Implicit Rule Learning via GPT-4}

\name{Hyo Jeong Yun$^{1,2}$, Chanyoung Kim$^{3}$, Moonjeong Hahm$^{2}$, Kyuri Kim$^{4}$, Guijin Son$^{1,5*}$\thanks{* Corresponding author.}}

\address{MODULABS$^{1}$, Chung-ang University$^{2}$,  Konkuk University$^{3}$, Seoul Women's University$^{4}$, Yonsei University$^{5}$  \\
         dbsgywjd@cau.ac.kr, spthsrbwls123@yonsei.ac.kr\\
         }

\abstract{
Environmental, social, and governance (ESG) factors are widely adopted as higher investment return indicators. Accordingly, ongoing efforts are being made to automate ESG evaluation with language models to extract signals from massive web text easily. However, recent approaches suffer from a lack of training data, as rating agencies keep their evaluation metrics confidential. This paper investigates whether state-of-the-art language models like GPT-4 can be guided to align with unknown ESG evaluation criteria through strategies such as prompting, chain-of-thought reasoning, and dynamic in-context learning. We demonstrate the efficacy of these approaches by ranking 2nd in the Shared-Task ML-ESG-3 \textit{Impact Type} track for Korean without updating the model on the provided training data. We also explore how adjusting prompts impacts the ability of language models to address financial tasks leveraging smaller models with openly available weights. We observe longer general pre-training to correlate with enhanced performance in financial downstream tasks. Our findings showcase the potential of language models to navigate complex, subjective evaluation guidelines despite lacking explicit training examples, revealing opportunities for training-free solutions for financial downstream tasks.
 \\ \newline  }

\begin{document}

\maketitleabstract

\begin{figure*}[t]
    \centering
    \includegraphics[width=\textwidth]{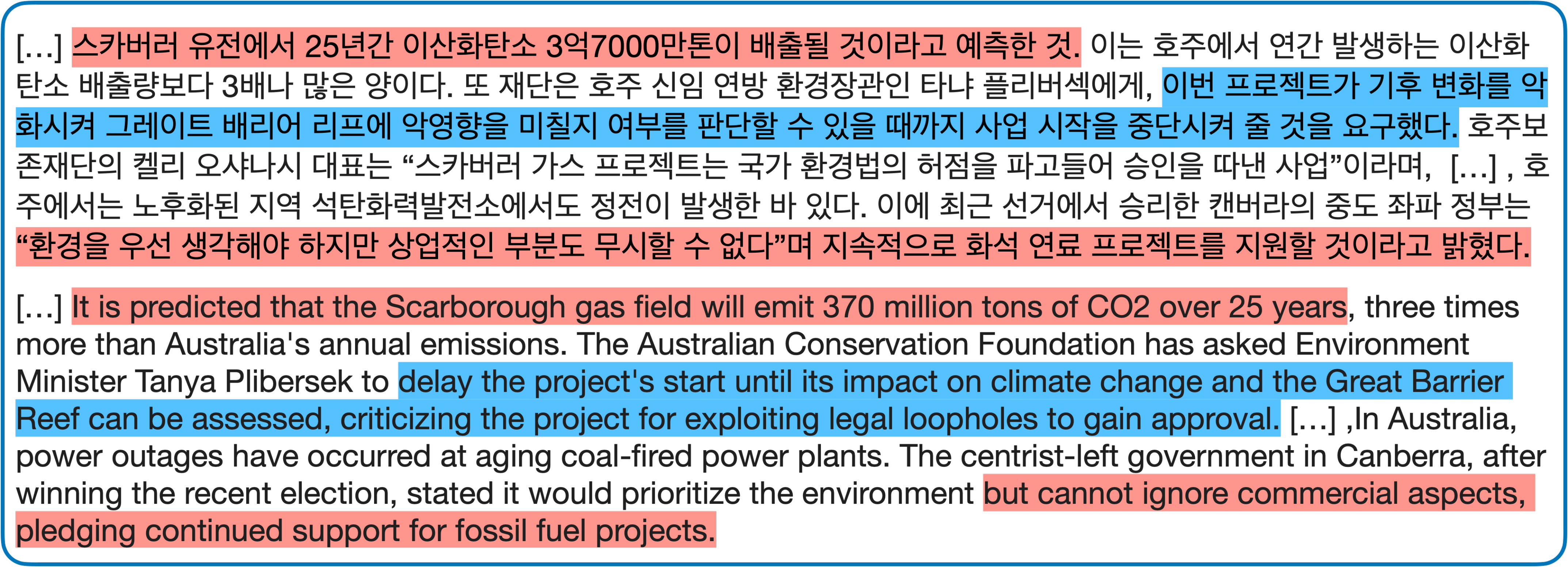}
    \caption{\footnotesize An example from the ML-ESG dataset. Sentences highlighted in red indicate negative implications for ESG, while those in blue denote positive ESG implications. The gold label for the ESG type of this text is "Opportunity." English translations are added for broader accessibility.}
    \label{fig:example}
\end{figure*}

\section{Introduction}
In recent years, there has been a noticeable increase in investors factoring environmental, social, and governance (ESG) considerations into their investment choices. Recent studies, through meta-analysis, have shown that improved ESG performance correlates with better corporate financial outcomes, potentially leading to higher investment returns~ \citep{cort2020esg, friede2015esg}. Assessing ESG performance involves nuanced analysis, and, as a result, the industry relies on rating agencies like MSCI\footnote{\url{https://www.msci.com}}, Sustainalytics\footnote{\url{https://www.sustainalytics.com}}, and Bloomberg\footnote{\url{https://www.bloomberg.com}} to evaluate and rank companies. Ongoing efforts to automate the ESG evaluation process exist, mainly through leveraging language models as substitutes for human analysts~\citep{mehra2022esgbert}. However, the specific methodologies used by each rating agency are not widely disclosed, leading to a lack of understanding of the detailed metrics necessary for evaluation. This closed nature presents challenges in training language models to accurately replicate the evaluation criteria these agencies employ. This presents challenges when training language models (especially the earlier versions like BERT~\citep{devlin2018bert}), as they rely heavily on explicit training data on the output distribution to accurately approximate the underlying function. Researchers have sought to enhance training datasets through synthetic data to address this issue~\citep{glenn-etal-2023-jetsons}. Nonetheless, several hurdles exist. First, the lack of transparency in the evaluation methodologies used by rating agencies, which often include subjective assessments, makes it difficult for researchers to generate realistic datasets. Moreover, the creation of large-scale, high-quality labeled datasets is resource-intensive. Manually annotating extensive text collections requires considerable time and skilled professionals. Furthermore, the accurate classification of sentences poses challenges due to the subjective nature of interpretation, which can vary even among experts~\citep{auzepy2023evaluating}. Finally, the rapid evolution of ESG criteria requires regular updates on the training dataset and retraining the model to align with changing investor expectations, emerging trends, and new reporting standards.

In this paper, we investigate whether state-of-the-art language models can be guided to align with unknown values (specifically, ESG evaluation standards) without learning from explicit training data. We employ multiple strategies, such as prompting, Chain-of-Thought reasoning~\citep{wei2022chain}, and dynamic in-context learning~\citep{dong2022survey} with \textit{GPT-4}~\citep{openai2024gpt4}, to participate in the Shared-Task ML-ESG-3 and rank second place in the \textit{Impact Type} track for Korean. Our findings underscore the efficacy of these strategies in approximating unknown guidelines, showcasing their potential in navigating the complexities of ESG criteria alignment. Furthermore, we extend our investigation to include two smaller models with publicly accessible weights, examining how slight modifications in prompts influence their performance and calibration. To the best of our knowledge, this study represents the first attempt to explore how adjustments in prompts can impact the ability of language models to address financial problems.

\section{Shared Task ML-ESG-3}
The Shared-Task ML-ESG-3 for Korean consists of two downstream tasks: \textit{Impact Type} and \textit{Impact Duration}. The \textit{Impact Type} task involves classifying given ESG news articles to one of \textit{Opportunity}, \textit{Risk}, or \textit{Cannot Distinguish}. The \textit{Impact Duration} task involves classifying the impact duration of a news article as one of \textit{Less than 2 years}, \textit{2 to 5 years}, or \textit{More than 5 years}. The dataset includes separate training and testing sets, with 800 Korean articles in the training set and 200 articles in the testing set.

\begin{table}[]
\centering
\fontsize{7}{9}\selectfont
\begin{tabular}{ccccc}
\toprule
\textbf{Category}   & \textbf{Opp.} & \textbf{Risk} & \textbf{Cannot Dist.}  & \textbf{Total.} \\
\midrule
Sustainable Econ.   & 160  & 57  & 41  & 258   \\
Corporate Govern.  & 134  & 31  & 40  & 205   \\
Env. \& Society   & 71  & 79 & 6    & 156 \\
Disclosure \& Eval. & 87  & 55 & 11   & 153 \\
ESG Life                 & 7   & 3  & 10   & 20 \\
Opinion                  & 3   & 4  & 1    & 8 \\
\midrule
Total                    & 462  & 229 & 109   & 800  \\
\bottomrule
\end{tabular}%
\caption{\footnotesize Statistics on the \textit{Impact Type} of Shared-Task ML-ESG-3 for Korean.}
\label{tab:stats_type}
\end{table}

\begin{table}[]
\centering
\fontsize{7}{9}\selectfont
\begin{tabular}{ccccc}
\toprule
\textbf{Category}  & \textbf{< 2 Yrs} &  \textbf{2-5 Yrs} & \textbf{> 5 Yrs} & \textbf{Total} \\
\midrule
Sustainable Econ.    & 101   & 54  & 103    & 258   \\
Corporate Govern.   & 137   & 36  & 32     & 205   \\
Env. \& Society   & 67   & 26   & 63   & 156    \\
Disclosure \& Eval. & 119   & 23  & 11   & 153    \\
ESG Life                 & 16    & 1   & 3    & 20     \\
Opinion                  & 6    & 2    & 0    & 8      \\
\midrule
Total           & 446   & 212  & 142  & 800 \\
\toprule
\end{tabular}%
\caption{\footnotesize Statistics on the \textit{Impact Duration} of Shared-Task ML-ESG-3 for Korean.}
\label{tab:stats_duration}
\end{table}

In Table~\ref{tab:stats_type} we illustrate the distribution of impact types across categories in the training dataset. We observe significant data imbalance across multiple columns. For instance, while the largest category, "Sustainable Economics" feature 258 samples, the smallest category "Opinions," only include eight. Furthermore, \textit{Opportunity} category comprises 462 entries, roughly four times the count of the \textit{Cannot Distinguish} category, which has 109 entries. The imbalance of data could potentially be attributed to either: 1) a sampling error arising from the small dataset size, or 2) the real-world distribution of ESG-related news being skewed, as press may be more reluctant to report negative issues due to associated risks. Regardless of the underlying cause, this imbalanced training set poses a critical challenge for traditional approaches to training language models, as they will inevitably learn skewed representations from the biased data distribution. Similar patterns can be found also for the \textit{Impact Duration} subset as shown in Table~\ref{tab:stats_duration}. The \textit{Less than 2 years} category is the largest with 446 entries, nearly three times more than the \textit{More than 5 years} category, which is the least represented with 142 entries. 

\section{Main Results}

In this section, we elaborate on our methodology(Section~\ref{sec:3.1}) and report observed performances (Section~\ref{sec:3.2}).

\subsection{Methodology}~\label{sec:3.1}

Predicting the ESG types and their impact duration from texts is a non-trivial task that traditionally relies on human experts. However, the criteria these experts use are mostly kept confidential. This ambiguity fence researchers from developing precise rules for LLMs to learn to perform such tasks. Accordingly, this leads to a question: \textbf{Can LLMs implicitly approximate unknown rules, without a comprehensive understanding of the task?} To address this question, we employ \textit{GPT-4}, a state-of-the-art language model. To align the model with the implicit rules we leverage the following approaches: 

\begin{figure}[t]
    \centering
    \includegraphics[width=\columnwidth]{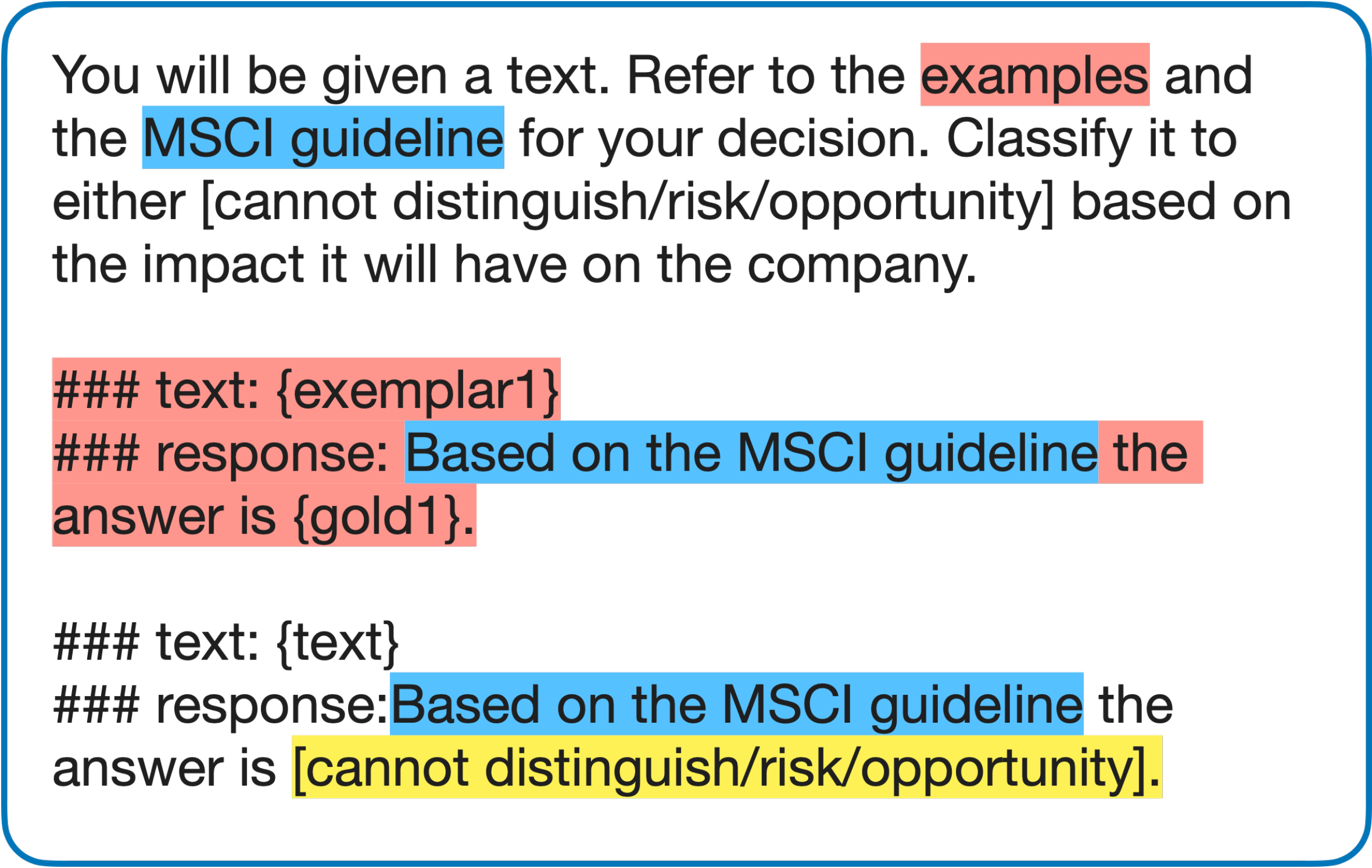}
    \caption{\footnotesize An example prompt with one examplar (highlighted in red) and prompts to follow the MSCI guidelines (highlghted in blue). We calculate the chance for the gold answer to follow "the answer is".}
    \label{fig:prompt}
\end{figure}

\paragraph{In-Context Learning}~\citep{dong2022survey}: In-context learning (ICL) is an approach where LLMs are provided with exemplars demonstrating the desired behavior. Instead of updating parameters through backpropagation, the model infers patterns from the examples and generalizes during inference. In our work, we dynamically alter the provided examples using the BM-25 algorithm. For a given input sample, we retrieve five relevant articles from the training set and provide them for ICL to the model during inference.

\paragraph{Chain-of-Thought}~\citep{wei2022chain}:Chain-of-thought guide models to generate a series of intermediate reasoning steps while solving a task. In an autoregressive structure, one forward pass is calculated per generated token; accordingly, allowing a model to generate intermediate reasoning allows it to leverage more forward passes as needed.

\paragraph{Prompt Engineering}~\citep{white2023prompt}:Prompt engineering involves creating prompts or prefixed to guide LLMs during inference. A prompt engineers the LLM to follow a desired behavior and output format. In this work, we prompt the language model to follow the MSCI guidelines for classification.

\begin{figure}[h]
    \centering
    \includegraphics[width=\columnwidth]{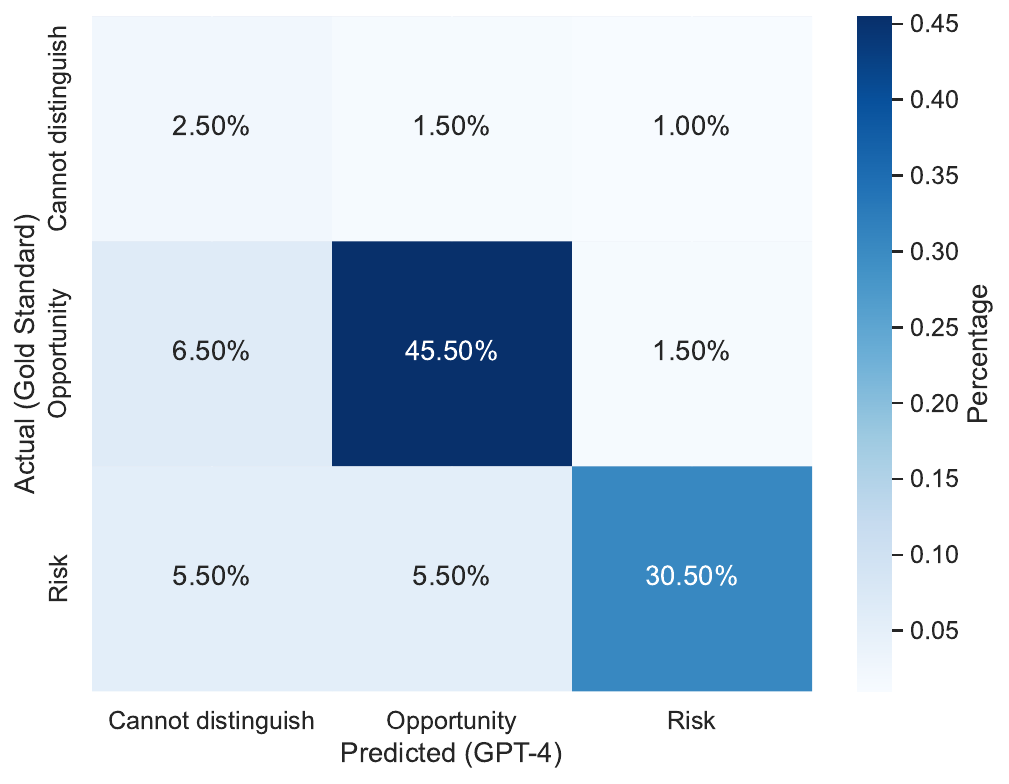}
    \caption{\footnotesize A confusion matrix analyzing the performance of \textit{GPT-4} on the \textit{Impact Type} subset.}
    \label{fig:cm_type}
\end{figure}

\begin{figure}[h]
    \centering
    \includegraphics[width=\columnwidth]{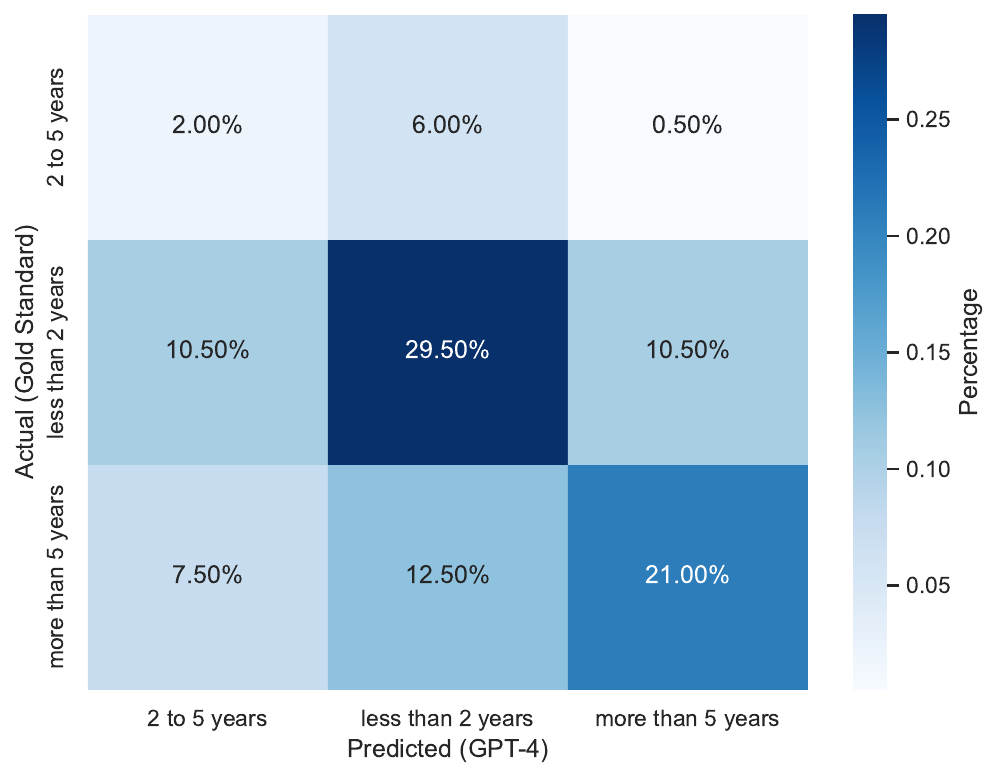}
    \caption{\footnotesize A confusion matrix analyzing the performance of \textit{GPT-4} on the \textit{Impact Duration} subset.}
    \label{fig:cm_duration}
\end{figure}

\subsection{Evaluation Results}~\label{sec:3.2}

\begin{table*}[t]
\centering
\fontsize{8.5}{11}\selectfont
\begin{tabular}{llcccc}
\toprule
\multicolumn{1}{c}{\textbf{Task}} & \multicolumn{1}{c}{\textbf{Model}} & \textbf{Min} & \textbf{Max} & \textbf{Mean} & \textbf{$\Delta$ (Max - Min)} \\
\midrule
Impact Duration & EEVE-Korean-10.8B & 38.0 & 48.5 & 44.9 & 10.5 \\
Impact Type & EEVE-Korean-10.8B & 35.0 & 55.5 & 48.9 & 20.5 \\
Impact Duration & Yi-Ko-6B & 44.0 & 51.5 & 47.9 & 7.5 \\
Impact Type & Yi-Ko-6B & 59.0 & 65.5 & 63.2 & 6.5 \\ 
\bottomrule
\end{tabular}
\caption{\footnotesize Performance summary of \textit{Yi-Ko-6B} and \textit{EEVE-Korean-10.8B} with ten different prompts. We report the accuracy (\%) of each models.}
\label{tab:calibration}
\end{table*}

\begin{table}[h]
\centering
\fontsize{8}{11}\selectfont
\begin{tabular}{ccc}
\toprule
\textbf{Submission} & \textbf{Impact Type} & \textbf{Impact Duration} \\
\midrule
Ours & \underline{76.13} & 43.98 \\
\midrule
3idiots\_3 & \textbf{79.85} & \underline{61.54} \\
Jetsons\_1 & - & \textbf{66.24} \\
Tredence\_2 & 75.95 & 58.18 \\ 
\bottomrule
\end{tabular}
\caption{\footnotesize Performance of selected models. The highest-scoring model is highlighted in \textbf{bold}, and second-highest is \underline{underlined}.}
\label{tab:main}
\end{table}

Table~\ref{tab:main} showcases the performance of selected models on the Korean subset for the Shared Task ML-ESG-3. Notably, our approach, which utilizes 5-shot exemplars and prompt engineering based on MSCI guidelines, ranks second in \textit{Impact Type} classification. However, it falls short in accurately predicting \textit{Impact Duration}. An initial analysis of the outputs, presented in Figures~\ref{fig:cm_type} and \ref{fig:cm_duration}, reveals a tendency of \textit{GPT-4} to incorrectly classify impact durations as \textit{less than 2 years}. Further qualitative examination shows that articles containing multiple perspectives and events often mislead the model. This observation is consistent with findings that LLMs struggle with comprehending and referencing longer text inputs~\citep{levy2024same}. An example highlighting an instance with multiple implications is provided in Figure~\ref{fig:example}. Despite the challenges, SOTA LLMs like GPT-4 demonstrate a remarkable ability to implicitly identify patterns, surpassing traditional performance methods without requiring specific training.

\section{Calibration}~\label{sec:4}

For a model's decisions to be considered trustworthy, they must be well-calibrated; this means that its confidence levels should accurately reflect the true likelihood of its predictions being correct. In this section, we will explore how various approaches influence models' calibration and accuracy.

\subsection{Experimental Settings}

\paragraph{Models}~Unfortunately, the \textit{GPT-4} API does not provide enough information for the intended analysis. Therefore, we choose to use \textit{Yi-Ko-6B}~\citep{yiko} and \textit{EEVE-Korean-10.8B}~\citep{kim2024efficient} two pre-trained models with fewer than 14 billion parameters that demonstrate the highest performance on the KMMLU~\citep{son2024kmmlu} benchmark. See Appendix~\ref{ap.1} for further details on the models.

\paragraph{Evaluation}~We evaluate ten distinct approaches, varying the number of in-context exemplars, the order of these exemplars, and the prompts themselves. See Appendix~\ref{ap.1} for an explanation of each approach. For each approach, we append "The answer is" to a query and calculate the likelihood of each option following the query. Figure~\ref{fig:prompt} provides an example of the query format. 

\begin{figure}[h]
    \centering
    \includegraphics[width=\columnwidth]{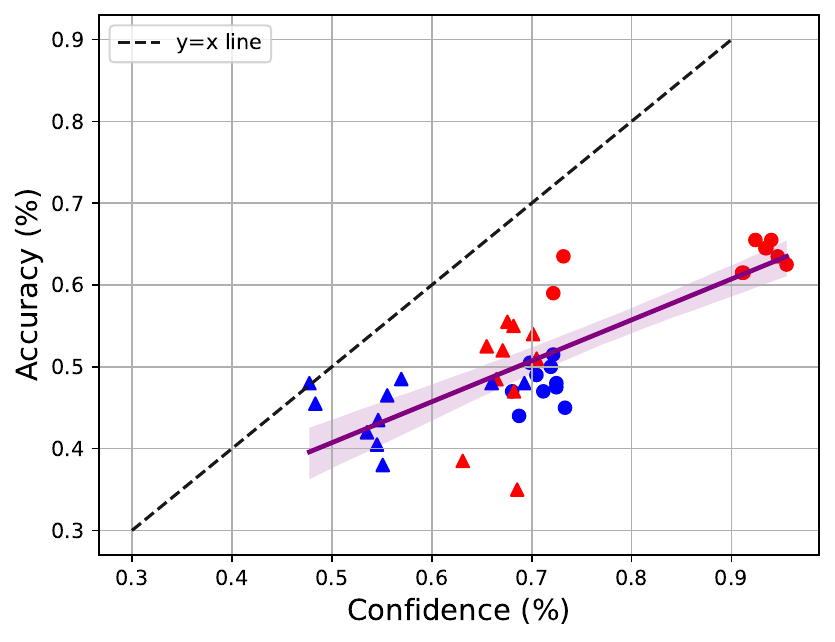}
    \caption{\footnotesize Relationship between accuracy and confidence of Yi-Ko-6B (circle) and EEVE-Korean-10.8B (triangle) for both subsets.(Red for 'Impact Type' and blue for 'Impact Duration'). Regression analysis exhibits a slope of 0.50.}
    \label{fig:calibration}
\end{figure}

\subsection{Analysis}

In Figure~\ref{fig:calibration}, we provide an overview of the calibration of models by testing how well the average confidence estimates the accuracy for each prompt. Surprisingly, both model appears to be well-calibrated, with a regression analysis exhibiting a slope of 0.5. In Table~\ref{tab:calibration}, we observe that \textit{Yi-Ko-6B} outperforms \textit{EEVE-Korean-10.8B} in both average and maximum scores. Additionally, \textit{Yi-Ko-6B} exhibits a smaller delta, indicating greater robustness to prompt variations. This increased robustness may stem from extended continual pre-training, which is consistent with recent studies suggesting that the ICL capabilities of models are enhanced by encountering parallel structures in the training corpora~\citep{chen2024parallel}. Extended continual pre-training in Korean likely increases the model's exposure to parallel structures, thus improving its ability to capture implicit patterns robustly. Our analysis indicates that smaller, publicly available models can also effectively identify implicit patterns in ESG classification without prior training. Without needing task-specific fine-tuning, general pre-training seems to improve their robustness and overall performance.

\section{Conclusion}

In this work, we adopt multiple prompting, chain-of-thought reasoning, and in-context learning strategies to guide \textit{GPT-4} in solving ESG classification tasks. We rank second in the Korean subset for Shared Task ML-ESG-3 in \textit{Impact Type} prediction. Furthermore, we adopt open models to explain their calibration and robustness to different prompting strategies. The longer general pre-training correlates with enhanced performance in financial downstream tasks. While our work has been limited to the Korean language, we believe it will be equally applicable in different languages, especially in English, and leave for future works. 

\section*{Acknowledgements}
This research was supported by Brian Impact Foundation, a non-profit organization dedicated to the advancement of science and technology for all.
 
\vspace{5mm}

\nocite{*}
\section{Bibliographical References}\label{sec:reference}

\bibliographystyle{lrec-coling2024-natbib}
\bibliography{krxb}

\bibliographystylelanguageresource{lrec-coling2024-natbib}
\bibliographylanguageresource{languageresource}

\appendix

\onecolumn

\section{Additional details for Section~\ref{sec:4}}~\label{ap.1}

\subsection{Adopted Models}

We adopt the following models with openly-available weights for analysis in Section~\ref{sec:4}. Due to hardware limitations, all models are used in 4-bit quantization.

\begin{enumerate}
    \item \textit{EEVE-Korean-10.8B}~\citep{kim2024efficient}:  A Korean vocabulary-extended ver sion of \textit{SOLAR-10.7B}~\citep{kim2023solar} that has undergone continual pre-training on a total of 3.2M documents (or, 3.2B tokens).
    \item \textit{Yi-Ko-6B}~\citep{yiko}:  A Korean vocabulary-extended version of Yi-6B~\citep{yi} that has undergone continual pre-training on 60B tokens. 
\end{enumerate}

\subsection{Prompts}
In Table~\ref{tab:prompts}, we provide an overview of the ten prompts used for analysis in Section~\ref{sec:4}.

\begin{table*}[h]
\centering
\fontsize{8}{11}\selectfont
\begin{tabular}{cccc}
\toprule
\textbf{Prompt Name} & \textbf{\# of In-Context Exemplars} & \textbf{Order of Exemplars} & \textbf{Prompted to follow MSCI Guidelines} \\
\midrule
1-shot-standard\_order-msci & 1 & Similar First & O \\
1-shot-standard\_order-standard & 1 & Similar First & X \\
3-shot-reverse\_order-msci & 3 & Similar Last & O \\
3-shot-reverse\_order-standard & 3 & Similar Last & X \\
3-shot-standard\_order-msci & 3 & Similar First & O \\
3-shot-standard\_order-standard & 3 & Similar First & X \\
5-shot-reverse\_order-msci & 5 & Similar Last & O \\
5-shot-reverse\_order-standard & 5 & Similar Last & X \\
5-shot-standard\_order-msci & 5 & Similar First & O \\
5-shot-standard\_order-standard & 5 & Similar First & X \\
\bottomrule
\end{tabular}
\caption{Entire list of prompt settins used in Section~\ref{sec:4}.}
\label{tab:prompts}
\end{table*}

\subsection{Performance Details}

In Tables~\ref{tab:yiko} and \ref{tab:yiko} we present the detailed per prompt perfomrnace for each models. 

\begin{table}[h]
\centering
\fontsize{9}{12}\selectfont
\begin{tabular}{ccccc}
\toprule
\textbf{Prompt} & \textbf{Accuracy} & \textbf{Confidence} & \textbf{Model} & \textbf{Task} \\
\midrule
1-shot-standard\_order-msci\_simple & 0.635 & 0.731760 & Yi-Ko-6B & Impact Type \\
1-shot-standard\_order-standard & 0.590 & 0.721608 & Yi-Ko-6B & Impact Type \\
3-shot-reverse\_order-msci\_simple & 0.625 & 0.955045 & Yi-Ko-6B & Impact Type \\
3-shot-reverse\_order-standard & 0.635 & 0.946185 & Yi-Ko-6B & Impact Type \\
3-shot-standard\_order-msci\_simple & 0.645 & 0.933864 & Yi-Ko-6B & Impact Type \\
3-shot-standard\_order-standard & 0.655 & 0.923851 & Yi-Ko-6B & Impact Type \\
5-shot-reverse\_order-msci\_simple & 0.645 & 0.934855 & Yi-Ko-6B & Impact Type \\
5-shot-reverse\_order-standard & 0.655 & 0.939728 & Yi-Ko-6B & Impact Type \\
5-shot-standard\_order-msci\_simple & 0.615 & 0.910514 & Yi-Ko-6B & Impact Type \\
5-shot-standard\_order-standard & 0.615 & 0.912037 & Yi-Ko-6B & Impact Type \\
\midrule
1-shot-standard\_order-msci & 0.505 & 0.698373 & Yi-Ko-6B & Impact Duration \\
1-shot-standard\_order-standard & 0.500 & 0.719090 & Yi-Ko-6B & Impact Duration \\
3-shot-reverse\_order-msci & 0.470 & 0.680418 & Yi-Ko-6B & Impact Duration \\
3-shot-reverse\_order-standard & 0.490 & 0.704762 & Yi-Ko-6B & Impact Duration \\
3-shot-standard\_order-msci & 0.475 & 0.724632 & Yi-Ko-6B & Impact Duration \\
3-shot-standard\_order-standard & 0.515 & 0.721509 & Yi-Ko-6B & Impact Duration \\
5-shot-reverse\_order-msci & 0.440 & 0.687383 & Yi-Ko-6B & Impact Duration \\
5-shot-reverse\_order-standard & 0.470 & 0.711635 & Yi-Ko-6B & Impact Duration \\
5-shot-standard\_order-msci & 0.450 & 0.733333 & Yi-Ko-6B & Impact Duration \\
5-shot-standard\_order-standard & 0.480 & 0.724686 & Yi-Ko-6B & Impact Duration \\
\bottomrule
\end{tabular}
\caption{\footnotesize Detailed performance of \textit{Yi-Ko-6B} on different prompts.}
\label{tab:yiko}
\end{table}

\begin{table}[]
\centering
\fontsize{9}{11}\selectfont
\begin{tabular}{ccccc}
\toprule
\textbf{Prompt} & \textbf{Accuracy} & \textbf{Confidence} & \textbf{Model} & \textbf{Task} \\
\midrule
1-shot-standard\_order-msci\_simple & 0.35 & 0.685465 & EEVE-Korean-10.8B & Impact Type \\
1-shot-standard\_order-standard & 0.385 & 0.630959 & EEVE-Korean-10.8B & Impact Type \\
3-shot-reverse\_order-msci\_simple & 0.525 & 0.654941 & EEVE-Korean-10.8B & Impact Type \\
3-shot-reverse\_order-standard & 0.54 & 0.701319 & EEVE-Korean-10.8B & Impact Type \\
3-shot-standard\_order-msci\_simple & 0.485 & 0.664646 & EEVE-Korean-10.8B & Impact Type \\
3-shot-standard\_order-standard & 0.55 & 0.681784 & EEVE-Korean-10.8B & Impact Type \\
5-shot-reverse\_order-msci\_simple & 0.51 & 0.704919 & EEVE-Korean-10.8B & Impact Type \\
5-shot-reverse\_order-standard & 0.555 & 0.675689 & EEVE-Korean-10.8B & Impact Type \\
5-shot-standard\_order-msci\_simple & 0.47 & 0.682284 & EEVE-Korean-10.8B & Impact Type \\
5-shot-standard\_order-standard & 0.52 & 0.670969 & EEVE-Korean-10.8B & Impact Type \\
\midrule
1-shot-standard\_order-msci & 0.48 & 0.659873 & EEVE-Korean-10.8B & Impact Duration \\
1-shot-standard\_order-standard & 0.48 & 0.692712 & EEVE-Korean-10.8B & Impact Duration \\
3-shot-reverse\_order-msci & 0.435 & 0.546392 & EEVE-Korean-10.8B & Impact Duration \\
3-shot-reverse\_order-standard & 0.465 & 0.555405 & EEVE-Korean-10.8B & Impact Duration \\
3-shot-standard\_order-msci & 0.42 & 0.535136 & EEVE-Korean-10.8B & Impact Duration \\
3-shot-standard\_order-standard & 0.485 & 0.569464 & EEVE-Korean-10.8B & Impact Duration \\
5-shot-reverse\_order-msci & 0.405 & 0.545175 & EEVE-Korean-10.8B & Impact Duration \\
5-shot-reverse\_order-standard & 0.48 & 0.477536 & EEVE-Korean-10.8B & Impact Duration \\
5-shot-standard\_order-msci & 0.38 & 0.55096 & EEVE-Korean-10.8B & Impact Duration \\
5-shot-standard\_order-standard & 0.455 & 0.483521 & EEVE-Korean-10.8B & Impact Duration \\ 
\bottomrule
\end{tabular}
\caption{\footnotesize Detailed performance of \textit{EEVE-Korean-10.8B} on different prompts.}
\label{tab:eeve}
\end{table}

\end{document}